\pdfoutput=1
\documentclass[11pt]{article}

\usepackage[]{ACL2023}
\usepackage{times}
\usepackage{latexsym}

\usepackage{adjustbox}
\usepackage{multirow}
\usepackage{amsmath}  % for equation

\usepackage{booktabs} % for three-line table

\usepackage[T1]{fontenc}
\usepackage[utf8]{inputenc}

\usepackage{microtype}

\usepackage{inconsolata}

\title{BranchNorm: Robustly Scaling Extremely Deep Transformers}

\author{
  Yijin Liu\thanks{\quad Equal contributions.},~
  Xianfeng Zeng\footnotemark[1],~ 
  Fandong Meng\thanks{\quad  Corresponding author.} ~
  and Jie Zhou \\
  Pattern Recognition Center, WeChat AI, Tencent Inc, China \\
  \texttt{\{yijinliu, xianfzeng, fandongmeng, withtomzhou\}@tencent.com} \\
}

\begin{document}
\maketitle

\begin{abstract}
Recently, DeepNorm scales Transformers into extremely deep ({\em i.e.,} 1000 layers) and reveals the promising potential of deep scaling. 
To stabilize the training of deep models, DeepNorm~\citep{deepnet_2022} attempts to constrain the model update to a constant value. 
Although applying such a constraint can benefit the early stage of model training, it may lead to undertrained models during the whole training procedure. 
In this paper, we propose BranchNorm, which dynamically rescales the non-residual branch of Transformer in accordance with the training period. 
BranchNorm not only theoretically stabilizes the training with smooth gradient norms at the early stage, but also encourages better convergence in the subsequent training stage. 
Experiment results on multiple translation tasks demonstrate that BranchNorm achieves a better trade-off between training stability and converge performance.
% ({\em i.e.,} 0.6 BLEU improvement over DeepNorm on the WMT2014 En-Fr dataset).
\end{abstract}

\section{Introduction}
In recent years, Transformers~\citep{transformer_2017} have been developed rapidly and achieved state-of-the-art (SOTA) performance on a wide range of tasks. Meanwhile, the model capacity gets substantially expanded by widening the model dimension~\citep{BERT_2019,roberta_2019,xlmr_xl_2021,m6_10t_2021,megatron_2022}. 
Given that deep neural models learn feature representations with multiple layers of abstraction~\citep{deep_learning_2015}, it is more attractive to increase model capacity by scaling depths than widths.
Unfortunately, due to the training instability of Transformers, the depths of these SOTA models are still relatively shallow~\citep{scaling_laws_2020,scaling_laws_2022}.

\begin{figure}[t!]
\begin{center}
     \scalebox{0.5}{
      \includegraphics[width=1\textwidth]{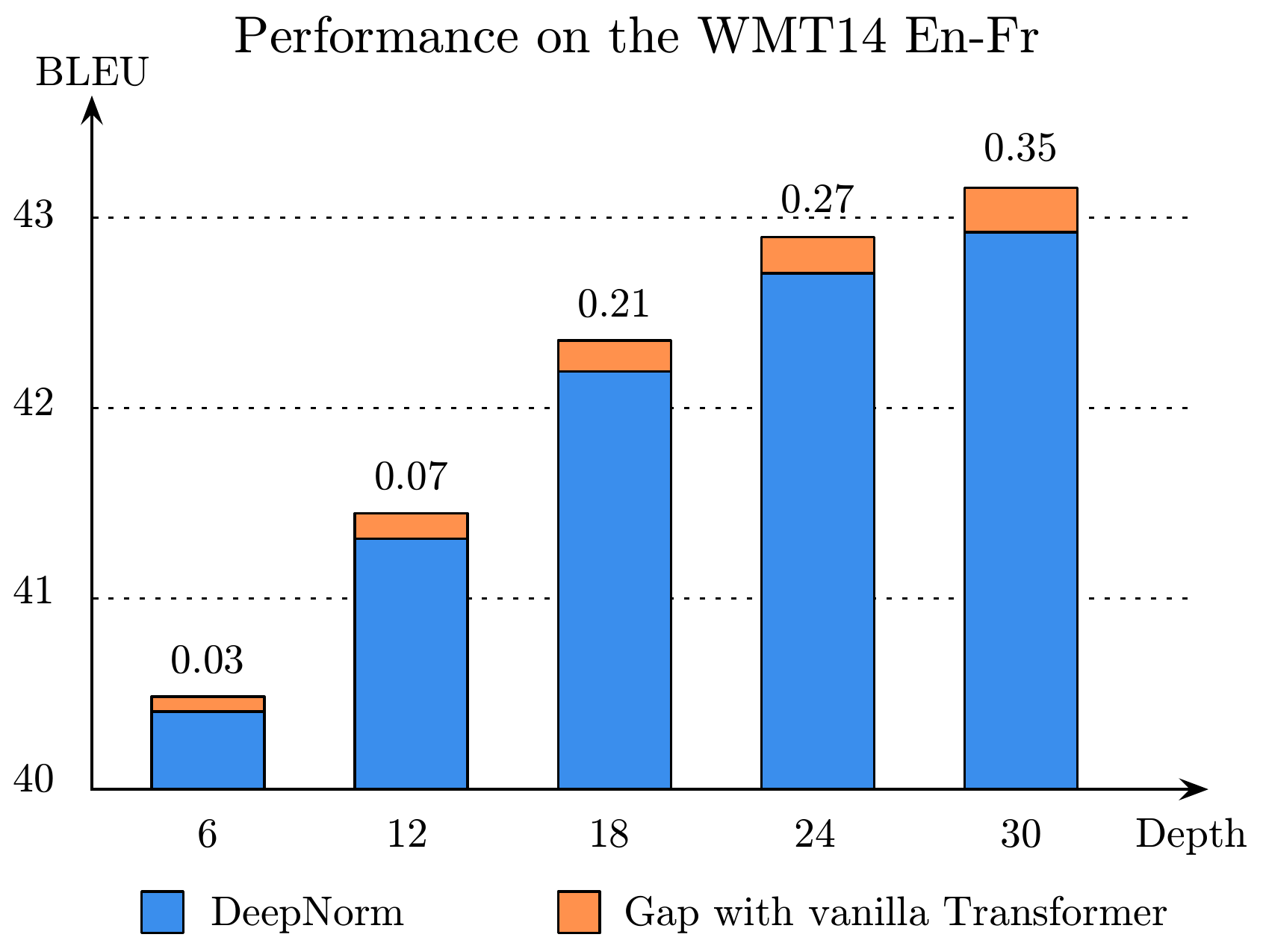}
      } 
      \caption{
      BLEU(\%) scores on the WMT2014 En-Fr dataset after models fully converge. `Gap' refers to the performance decline observed after applying DeepNorm on the vanilla Transformer.
      } 
      \label{fig:gain_hist}  
 \end{center} 
\end{figure}

To stabilize the training of Transformers, there have been various efforts on better architectures~\citep{DLCL_2019,Normformer2021,fundation_transformer}, or the implementation of proper initialization~\citep{BiaoZhang2019ImprovingDT,better_init_2020,deepnet_2022}. 
Among them, the most representative approach is DeepNorm~\citep{deepnet_2022}, which first scales Transformers to 1000 layers and significantly outperforms existing shallow counterparts.

Specifically, DeepNorm aims to constrain the model update to a constant level by upweighting the residual connections in Transformer and reducing the variance of parameter initialization. 
As a result, the stability of Transformers is improved in the early training stage.
However, in the subsequent training stage, the limitation of the magnitude of parameter updates imposed by DeepNorm may ultimately yield undertrained models.
% DeepNorm limits the magnitude of parameter updates and may ultimately yield undertrained models.
To verify the above conjecture, we first conduct experiments on shallow Transformers to guarantee 
convergences. As shown in Figure~\ref{fig:gain_hist}, it is observed that DeepNorm brings a certain degree of performance decline on vanilla Transformers, and this issue tends to get worse when models get deeper.

To address the above issue, we propose a simple yet effective approach to robustly scale extremely deep Transformers, named BranchNorm. 
Specifically, the non-residual branch\footnote{Note that we name the residual connections in Transformer as `residual branch' and the other branch as `non-residual branch' in this paper.} of the Transformer is dynamically rescaled  in accordance with the training period. 
In the early stage of model training, BranchNorm theoretically stabilizes the training with
smooth gradient norms. While in the subsequent training stage, BranchNorm progressively degenerates into vanilla Post-LayerNorm ({\em i.e.,} Post-LN) to promote better convergence.
Experiments on a wide range of translation tasks show that BranchNorm brings consistent improvement over DeepNorm, and effectively alleviates the above undertrained issue.
Moreover, BranchNorm performs more robustly on some key hyperparameters ({\em e.g.,} warmup) than DeepNorm, which makes it likely to be a portable alternative for scaling extremely deep Transformers.

The contributions of this paper can be summarized as follows:
% \footnote{The codes will be made publicly once acceptance.}:
% available at https://github.com/Adaxry/BranchNorm}:
\begin{itemize}
    \item We propose a simple yet effective normalization approach, named BranchNorm, to stabilize the training of extremely deep Transformers.
    \item BranchNorm achieves a better trade-off between training stability and converges performance on a wide range of translation tasks.
    \item BranchNorm is demonstrated to alleviate the problem of parameter redundancy in extremely deep models, from the perspective of representing similarity and sparsity of activation functions.
\end{itemize}

\begin{figure}[t!]
\begin{center}
     \scalebox{0.45}{
      \includegraphics[width=1\textwidth]{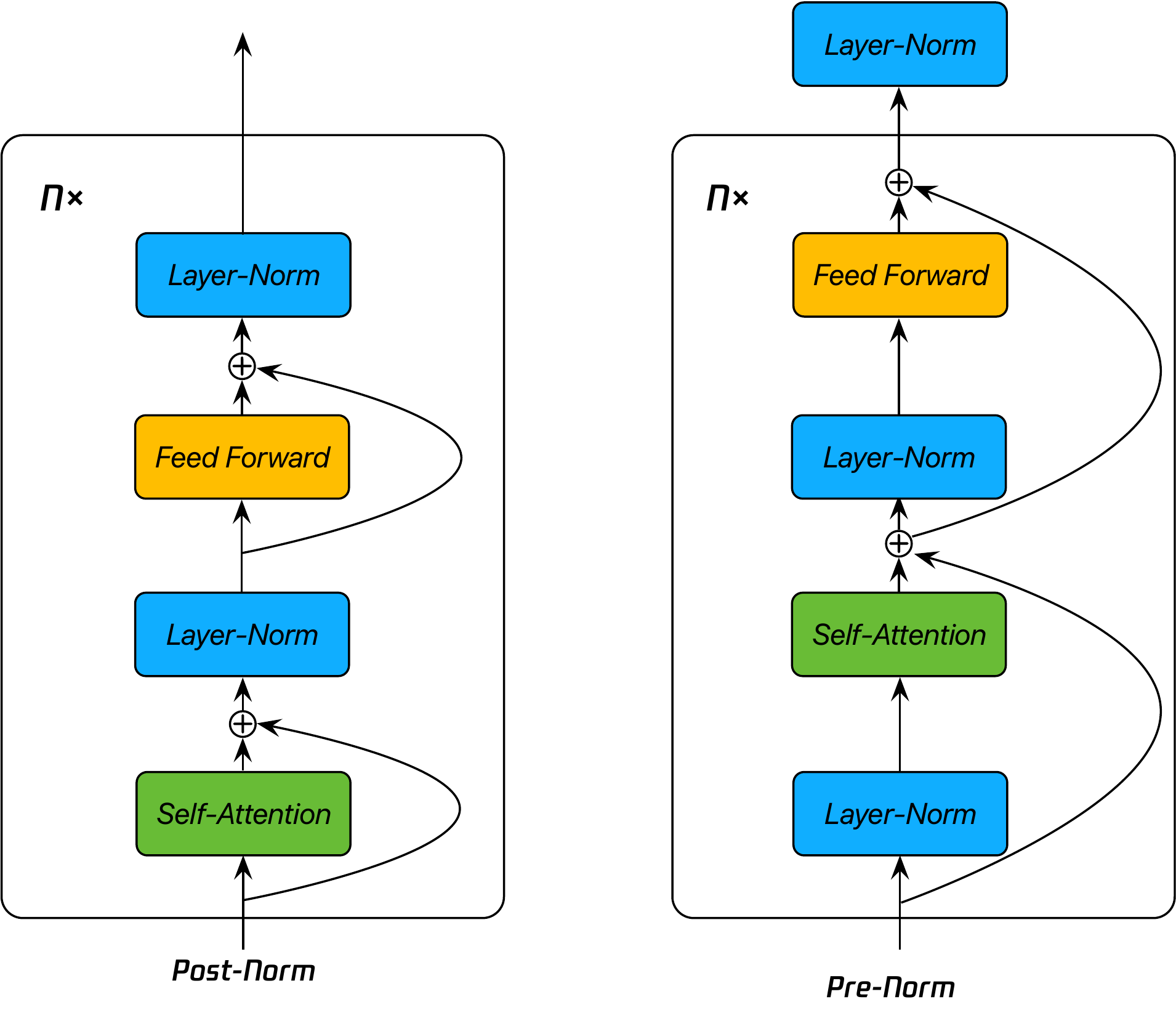}
      } 
      \caption{
      The architectures of Pre-Norm ({\em i.e.,} Pre-LN) and Post-Norm ({\em i.e.,} Post-LN) Transformers.
      } 
      \label{fig:prenorm_Post-LN}  
 \end{center} 
\end{figure}

\section{Background}
In this section, we first provide a brief overview of the difference between Post-LN and Pre-LN, and subsequently introduce the approach of DeepNorm.

\paragraph{Post-LN and Pre-LN.}
Firstly, \citet{DLCL_2019,first_prenorm_2019} 
observe that the position of LayerNorm~\citep{ln_2016} has a significant effect on training stability, and propose the more stable Pre-LN variant when compared with the original Post-LN~\citep{transformer_2017}. 
An example of these two architectures is shown in Figure~\ref{fig:prenorm_Post-LN}.
Subsequently, \citet{branch_dependency_2020} further analyze that Pre-LN may have an excessive reliance on its residual connections, 
% may depend too much on its residual connections and 
which inhibits the model from unleashing its full potential.
Motivated by the above observation, we base our approach on Post-LN in the remainder of our experiments. 

Formally, given the input of the sub-layer in the $l$-th sub-layer $x_{l}$, calculates the output $x_{l+1}$ is calculated by Post-LN as follows:
\begin{equation}
    x_{l+1} = LN(x_l + \mathcal{F}\left(x_l ; \theta_l\right))
\end{equation}
where $LN$ is an abbreviation for LayerNorm\footnote{For brevity, the two learnable parameters 
% $\gamma$ and $\beta$ 
in LayerNorm are omitted.}, $\mathcal{F}$  represents the function of the current sub-layer (attention or feed-forward) and $\theta_l$ denotes the corresponding parameters of the sub-layer.

\paragraph{DeepNorm.}
DeepNorm follows the Post-LN Transformer architecture and rescales the residual branch with a scalar factor $\alpha > 1$. Similarly, the $l$-th sub-layer is calculated by DeepNorm as follows:
\begin{equation}
x_{l+1} = LN(\alpha x_l + \mathcal{F}\left(x_l ; \theta_l\right))
\label{equ:deepnorm}
\end{equation}
In addition, DeepNorm reduces the variance of the initial parameters by scaling factor $\beta < 1$. Both the $\alpha$ and $\beta$ are functions of model depths, which are derived from the assumption of constant model update.
For a strandard Transformer with $N$-layer encoder and $M$-layer decoder, DeepNorm calculate $\alpha$ and $\beta$ as follows:

\begin{equation}
\begin{aligned}
\alpha_{encoder}&=0.81{(N^4M)}^{\frac{1}{16}}   \\ 
\beta_{encoder}&=0.87{(N^4M)}^{-\frac{1}{16}}   \\
\alpha_{decoder}&={(3M)}^{\frac{1}{4}} \\
\beta_{decoder}&={(12M)}^{-\frac{1}{4}}
\end{aligned}    
\end{equation}
Note that $\beta$ merely affects the model initialization, while $\alpha$ is used and fixed during the whole procedure.
Moreover, with the model getting deeper, DeepNorm assigns larger value of $\alpha$, which leads to the model outputs $x_{l+1}$ depend too much on the residual branch $\alpha x_{l}$ and thus ultimately yields the 
undertrained model parameters $\theta_l$ in Equation~(\ref{equ:deepnorm}).  

% ~\citep{normalized_activation_2022} propose to normalize activation functions to keep the variance of the gradient the same for all layers.

\section{Approaches}
In this section, we first analyze the instability of Post-LN from the perspective of gradient norm, then demonstrate how DeepNorm can alleviate the unbalanced gradients to a certain extent, and finally introduce our proposed method BranchNorm.

\subsection{Perspective of Gradient}
Unbalanced gradients are mainly responsible for the instability of Transformer\footnote{In recent years, there are also researchers questioning this point and providing different perspectives~\citep{branch_dependency_2020,deepnet_2022}.
Given that more explorations and discussions are needed to make it out, we still conduct analysis from the perspective of gradient norm in this paper.}~\citep{DLCL_2019,Normformer2021,gradinit_2021}, we firstly explore the relation between gradient and model depth following~\citet{DLCL_2019}. Given a Transformer with $L$ sub-layers and the training loss $\mathcal{E}$, the gradient for the $l$-th sub-layer is calculated by the chain rule\footnote{More detailed derivations are in Appendix~\ref{appendix:proof}.}:

\begin{equation}
% \resizebox{0.9\linewidth}{!}{$
\begin{aligned}
\frac{\partial \mathcal{E}}{\partial x_l}=
&\underbrace{\frac{\partial \mathcal{E}}{\partial x_L}}_{irreducible}
\times 
\underbrace{\prod_{k=l}^{L-1} \frac{\partial \mathrm{LN}\left(\mathcal{F}\left(x_k ; \theta_k\right)\right)}{\partial \mathcal{F}\left(x_k ; \theta_k\right)}}_{LN}
\times \\
&\underbrace{\prod_{k=l}^{L-1}\left(1+\frac{\partial \mathcal{F}\left(x_k ; \theta_k\right)}{\partial x_k}\right)}_{residual}
\end{aligned}
% $}
\label{equ:grad_postln}
\end{equation}
where the gradient consists of three terms and the last two items are multiplicative with respect to the number of model layers $L$. Once $L$ gets larger, the values of the last two items may become very large or very small, which can yield the gradient vanishing or exploding.
% the gradient of Post-LN will face the risk of vanishing or exploding.

Similarly, we analyze the gradient of DeepNorm based on Equation~(\ref{equ:deepnorm}), and get the gradient of the $l$-th sub-layer is calculated by:
\begin{equation}
\resizebox{1.0\linewidth}{!}{
$
\begin{aligned}
\frac{\partial \mathcal{E}}{\partial x_l}&=
\underbrace{\frac{\partial \mathcal{E}}{\partial x_L}}_{irreducible}
\times 
\underbrace{\prod_{k=l}^{L-1} \left( \frac{\partial \mathrm{LN}\left( \alpha x_k + \mathcal{F}\left(x_k ; \theta_l\right)  \right)}{\partial \left( \alpha  x_k + \mathcal{F}\left(x_k ; \theta_l\right) \right)} \right) }_{LN} \times  \\
& \ \ \ \ \ \ \ \ \ \underbrace{\prod_{k=l}^{L-1}\left(\alpha + \frac{\partial \mathcal{F}\left(x_k ; \theta_k\right)}{\partial x_k} \right) }_{residual} \\
\end{aligned} 
\label{equ:grad_deepnorm}
$
}
\end{equation}
In DeepNorm, $\alpha$ increases with model depth and helps with training stability. Theoretically, $\alpha$ can go to infinity to represent the upper bound of DeepNorm's stability.
Here, we introduce this assumption to simplify the derivation: If $\alpha$ get large enough, {\em i.e.,} $\alpha \to \infty$, the LN item can be approximated as $\prod_{k=l}^{L-1} \frac{\partial \mathrm{LN}\left(\alpha x_k \right) }{\partial \left(\alpha x_k \right)}$, and the residual item can be approximated as $\prod_{k=l}^{L-1} \alpha $, we put them into Equation (\ref{equ:grad_deepnorm}) and can simplify it as follows: 
\begin{equation}
\resizebox{0.98\linewidth}{!}{$
\begin{aligned}
\frac{\partial \mathcal{E}}{\partial x_l}& 
\approx \underbrace{\frac{\partial \mathcal{E}}{\partial x_L}}_{irreducible}
\times 
\underbrace{\prod_{k=l}^{L-1} \left(\frac{\partial \mathrm{LN}\left(\alpha x_k\right)}{\partial \left( \alpha x_k \right)}\right) }_{LN}
\times \underbrace{\prod_{k=l}^{L-1} \alpha  }_{residual} 
\\
&=\underbrace{\frac{\partial \mathcal{E}}{\partial x_L}}_{irreducible}
\times 
\underbrace{\prod_{k=l}^{L-1} \left(\frac{\partial \mathrm{LN}\left(x_k\right)}{\partial x_k} \cdot  \frac{1}{\alpha} \right) }_{LN}
\times \underbrace{\prod_{k=l}^{L-1} \alpha  }_{residual} \\
&=\underbrace{\frac{\partial \mathcal{E}}{\partial x_L}}_{irreducible}
\times 
\underbrace{\prod_{k=l}^{L-1} \frac{\partial \mathrm{LN}\left(x_k\right)}{\partial x_k} }_{LN}   \ \ \ \ \ (\alpha \to \infty)
\end{aligned} 
\label{equ:grad_deepnorm_appro}
$}
\end{equation}
When compared to the gradient of Post-LN in Equation (\ref{equ:grad_postln}), DeepNorm can approximately eliminate the final residual item, and thus effectively mitigate the risk of gradient vanishing or exploding. 
Although a larger $\alpha$ in DeepNorm results in more stable gradients, it may come at the expense of final convergence performance, as mentioned above.
Considering that the unbalanced gradients generally occur during the early training stage, it may be more appropriate if $\alpha$ can be varied based on the training period.

\begin{figure}[t!]
\begin{center}
     \scalebox{0.48}{      \includegraphics[width=1\textwidth]{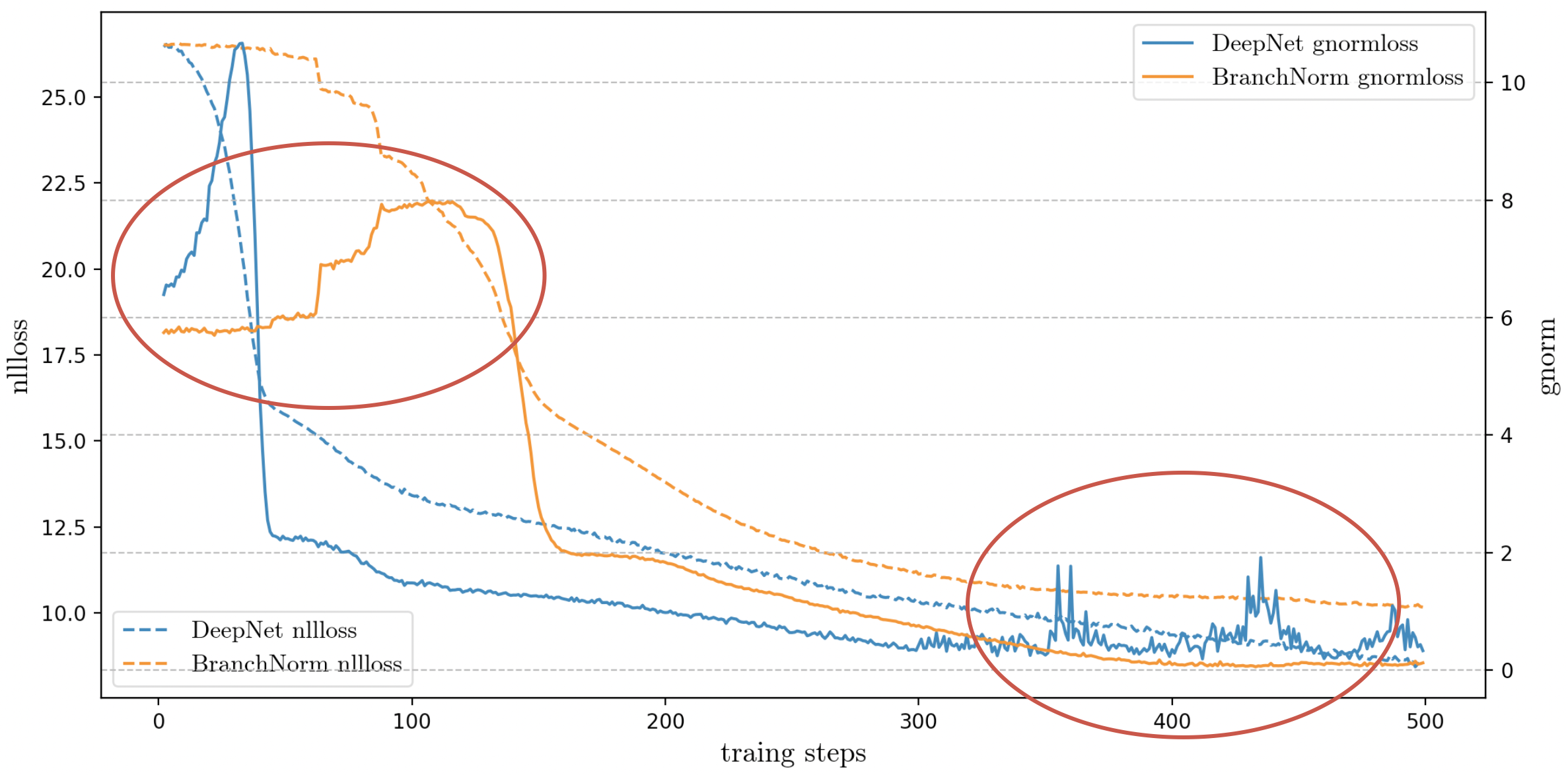}
      } 
      \caption{
      Gradient norm (solid line) and negative log likelihood loss (nllloss, dotted line) at the very beginning of training. 
      } 
      \label{fig:grad_loss}  
 \end{center}
\end{figure}

\subsection{BranchNorm}
In this section, we summarize the observations from previous sections and introduce BranchNorm:
\begin{equation}
x_{l+1} = LN(x_l + \alpha \mathcal{F}\left(x_l ; \theta_l\right))
\label{equ:branchnorm}
\end{equation}
It is analogous to the dual form of DeepNorm, but with two key differences:
First, BranchNorm utilizes a dynamic factor ({\em i.e.,} $\alpha$) that allows it to normalize gradients during the early training stage and gradually eliminate the negative effects of these normalizations during the later stage.
Second, the $\alpha$ of BranchNorm scales on the non-residual branches in Transformer, allowing for exactly normalize early stage's gradients without the strong assumptions of DeepNorm in Equation~(\ref{equ:grad_deepnorm}). 
Specifically, $\alpha$ in Equation~(\ref{equ:branchnorm}) is a simple factor with respect to the number of training step $t$. Here, we use a simple linear incremental approach:
\begin{equation}
\alpha_{t} = \min(1.0, t/T)
\label{equ:branchnorm_alpha}
\end{equation}
where $T$ is the maximum number of steps to conduct BranchNorm.
At the very beginning of training, $\alpha_{t}$ is approaching $0$, which means that the model approximates the constant transformation and gets updated with smooth gradients.
Following the analysis in the above section, we get the gradient of BranchNorm for the $l$-th sub-layer:

\begin{equation}
\begin{aligned}
\frac{\partial \mathcal{E}}{\partial x_l}&=
\underbrace{\frac{\partial \mathcal{E}}{\partial x_L}}_{irreducible}
\times 
\underbrace{\prod_{k=l}^{L-1}  \frac{\partial \mathrm{LN}\left(x_k + \alpha \mathcal{F}\left(x_k ; \theta_l\right) \right)}{\partial \left( x_k + \alpha \mathcal{F}\left(x_k ; \theta_l\right) \right)} }_{LN} 
\times \\ 
&\ \ \ \ \ \ \ \ \underbrace{\prod_{k=l}^{L-1} \left(1 + \alpha \frac{\partial \mathcal{F}\left(x_k ; \theta_l\right)  }{\partial x_k} \right)
 }_{residual}\\
&=\underbrace{\frac{\partial \mathcal{E}}{\partial x_L}}_{irreducible}
\times 
\underbrace{\prod_{k=l}^{L-1} \frac{\partial \mathrm{LN}\left(x_k\right)}{\partial x_k} }_{LN}  \ \ \ \ \ (\alpha = 0)
\\
\end{aligned} 
\end{equation}
At the very beginning of training,  BranchNorm can stabilize the gradient norm while DeepNorm requires a relatively strong assumption ($\alpha \to \infty$) in Equation~(\ref{equ:grad_deepnorm_appro}). 
Experimentally, as shown in Figure~\ref{fig:grad_loss}, we observe corresponding smoother gradients of BranchNorm at the very beginning of training.
Once the training step $t$ reaches the predefined maximum step $T$, BranchNorm degenerates to the vanilla Post-LN to achieve better convergence.
We further validate the hyperparameter insensitivity of BranchNorm in Section~\ref{sec:hyperpara_sensitivity}.

%%%%%%%%%%  WMT17 En-De table   %%%%%%%%%%
\begin{table*}[t]
\begin{center}
\scalebox{0.93}{
\begin{tabular}{l|c|ccccc}
\toprule
\textbf{Models} & \textbf{LN}  & \textbf{6L-6L} & \textbf{18L-18L} & \textbf{50L-50L} &
 \textbf{100L-100L} & \textbf{250L-250L} \\
\midrule
Vanilla Post-LN~\citep{transformer_2017} & Post & 28.1	&  \multicolumn{4}{c}{diverged} \\
 DS-Init~\citep{BiaoZhang2019ImprovingDT} & Post & 27.9 & \multicolumn{4}{c}{diverged}  \\
 Admin~\citep{branch_dependency_2020} & Post & 27.9 & 28.8 &  \multicolumn{3}{c}{diverged} \\
 \midrule
 ReZero~\citep{rezero2020} & No & 26.9 & \multicolumn{4}{c}{diverged} \\
 R-Fixup~\citep{HongyiZhang2019FixupIR} & No & 27.5 & 28.4 & 27.7 & diverged & diverged\\
 T-Fixup~\citep{better_init_2020} & No & 27.5 & 28.4 & 27.9	& diverged & diverged \\
 \midrule
 Vanilla Pre-LN~\citep{transformer_2017} & Pre & 27.0	& 28.1	& 28.0 & 27.4 & 27.5 \\
  DLCL~\citep{DLCL_2019} & Pre & 27.4 & 28.2 & diverged & 27.5 & 27.7
 \\
  NormFormer~\citep{Normformer2021} & Pre & 27.0 & 28.3 & 27.8 & diverged & diverged
 \\
 Sub-LN~\citep{fundation_transformer} $\dagger$ & Pre  & 27.5 & 28.3 & 28.7 & 27.7 & 27.9 \\
 DeepNorm~\citep{deepnet_2022} & Post & 27.8 & 28.8 & 29.0 & 28.9 & -- \\
 DeepNorm~\citep{deepnet_2022} $\dagger$  & Post & 28.6 & 29.1 & 29.7 & 29.3 & 29.0 \\
 \midrule
 % MixNorm (ours) & Post & \text{28.4} & \text{29.5} & \text{29.1} & \text{28.9} & \text{28.9} \\
 BranchNorm (ours) & Post & \textbf{29.3} & \textbf{30.3} & ~~\textbf{30.7}* & \textbf{29.8} & \textbf{29.6} \\
\bottomrule
\end{tabular}
}  % scale
\caption{BLEU scores (\%) on the WMT-17 En-De test set with depth-scaling. $\dagger$ indicates our reimplementations. $A$L-$B$L refers to a Transformer with $A$-layer encoder and $B$-layer decoder. `*' means BranchNorm is significantly better than DeepNorm with $p<0.03$.}
\label{tab:wmt17}
\end{center}
\end{table*}

%%%%%%%%%%  WMT14 En-Fr table   %%%%%%%%%%
\begin{table*}[t]
\begin{center}
\scalebox{0.93}{
\begin{tabular}{l|c|cccccc}
\toprule
\textbf{Models} & \textbf{LN}  & \textbf{6L-6L} & \textbf{18L-18L} & \textbf{50L-50L} &
 \textbf{100L-100L} & \textbf{250L-250L} & \textbf{500L-500L}  \\
\midrule
Vanilla Post-LN~(\citeyear{transformer_2017}) & Pre & 41.48 & 43.27 &  \multicolumn{4}{c}{diverged}\\
 Vanilla Pre-LN~(\citeyear{transformer_2017})  & Pre & 40.96 & 42.48	& 42.70 & 43.12 & 43.25  & 43.18 \\
DLCL~(\citeyear{DLCL_2019}) & Pre & 41.33 & 42.81 & 43.05 & \multicolumn{3}{c}{diverged} \\
% NormFormer~(\citeyear{Normformer2021}) & Pre & 27.0 & 28.3 & 27.8 & diverged & diverged & - \\
 Sub-LN~(\citeyear{fundation_transformer}) $\dagger$  & Pre  & 41.12 & 42.68 & 43.28 & 43.31 & 43.42  & 43.21 \\
 DeepNorm~(\citeyear{deepnet_2022}) $\dagger$  & Post & 41.47 & 42.92 & 43.79 & 43.93 & 43.87 & 43.67 \\
 \midrule
 MixNorm (ours) & Post & \textbf{41.96} & \text{43.34} & \text{43.81} & \text{43.91} & \text{43.73} 
 & \text{43.41} \\
 BranchNorm (ours) & Post & 41.67 & \textbf{43.53} & \textbf{43.89} & \textbf{44.20} & ~~\textbf{44.30}* 
 & \textbf{44.27} \\
 % BranchNorm (ours) & Post & \textbf{41.67} & \textbf{43.53} & \textbf{43.89} & \textbf{44.20} & \textbf{44.30} 
 % & \textbf{44.27} \\
\bottomrule
\end{tabular}
}  % scale
\caption{BLEU scores (\%) on the WMT-14 En-Fr test set with depth-scaling. $\dagger$ indicates our reimplementations. $A$L-$B$L refers to a Transformer with $A$-layer encoder and $B$-layer decoder. `*' means BranchNorm is significantly better than DeepNorm with $p<0.03$.}
\label{tab:wmt14_enfr}
\end{center}
\end{table*}

\section{Experiments and Results}
We conduct extensive experiments on both bilingual translation and multilingual translation tasks to verify our approach. 
In this section, we will describe our experimental settings and present results. 

\subsection{Datasets and Evaluations}
% 还需要描述仔细点，句子的量级，覆盖的语种数量
We use the standard WMT 2017 English-German (En-De), WMT 2014 English-French (En-Fr), and IWSLT 2014 German-English (De-En) datasets for the bilingual task, which is processed following the official scripts of fairseq\footnote{https://github.com/facebookresearch/fairseq}.
For the multilingual task, we conduct experiments on the OPUS-100~\citep{opus100} and MultiUN~\citep{multiun_2019} dataset and follow the corresponding processing in existing studies.

For evaluation, we set the beam size to 4 and the length penalty to 0.6 during inference. 
We use the \textit{multibleu.perl} to calculate cased sensitive BLEU scores for WMT 2017 En-De\footnote{For a rigorous comparison, we use the same test set with DeepNorm, {\em i.e.,} \textit{newstest2014.}} and WMT 2014 En-Fr.
Besides, we use sacreBLEU\footnote{https://github.com/mjpost/sacrebleu} to calculate cased sensitive BLEU scores for OPUS-100 and cased insensitive BLEU scores for MultiUN following~\citet{multiun_bleu}.
\begin{figure}[t!]
\begin{center}
     \scalebox{0.42}{
    \includegraphics[width=1\textwidth]{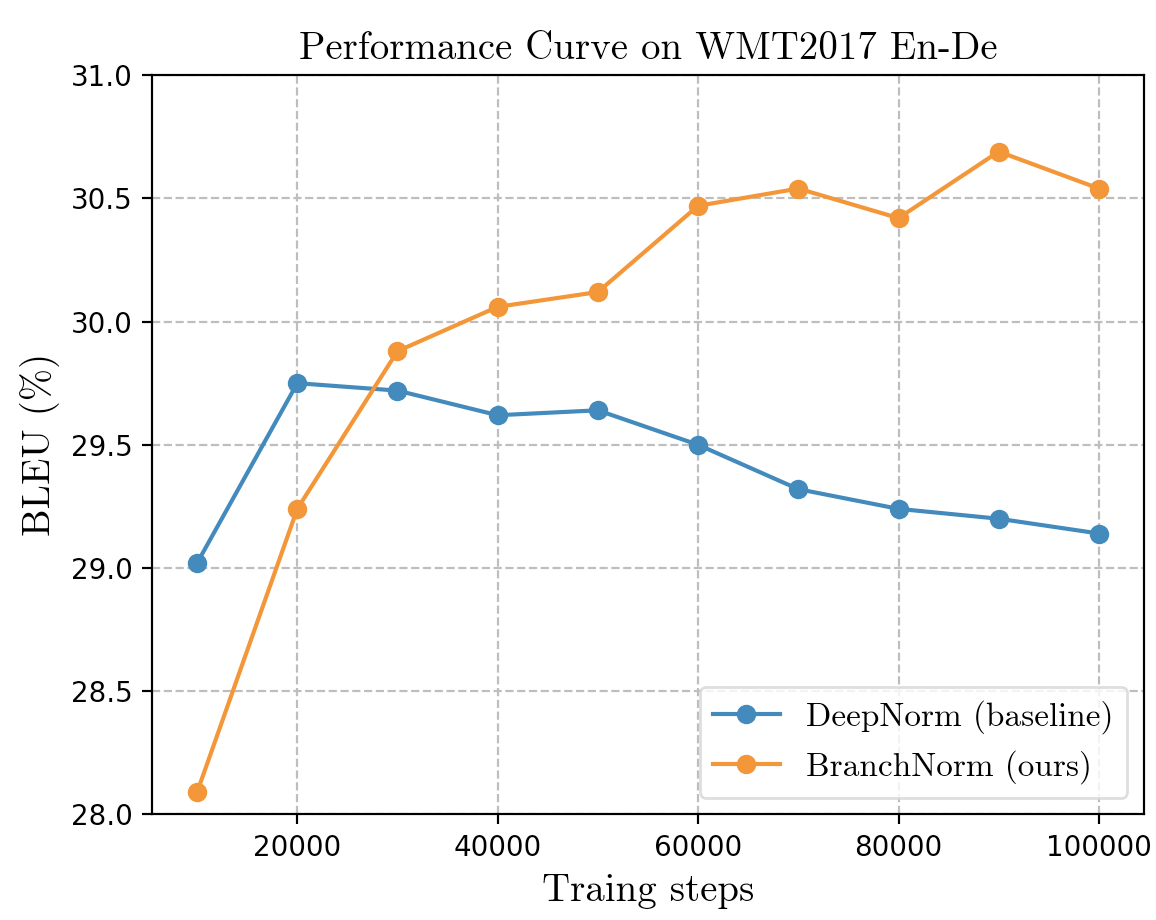}
      } 
      \caption{
      BLEU score curve of 50L-50L models on the WMT 2017 En-De with the increase of training steps.
      } 
      \label{fig:train_step_bleu}  
 \end{center}
 % \vspace{-20pt}
\end{figure}

\begin{figure*}[t!]
\begin{center}
     \scalebox{1}{
    \includegraphics[width=1\textwidth]{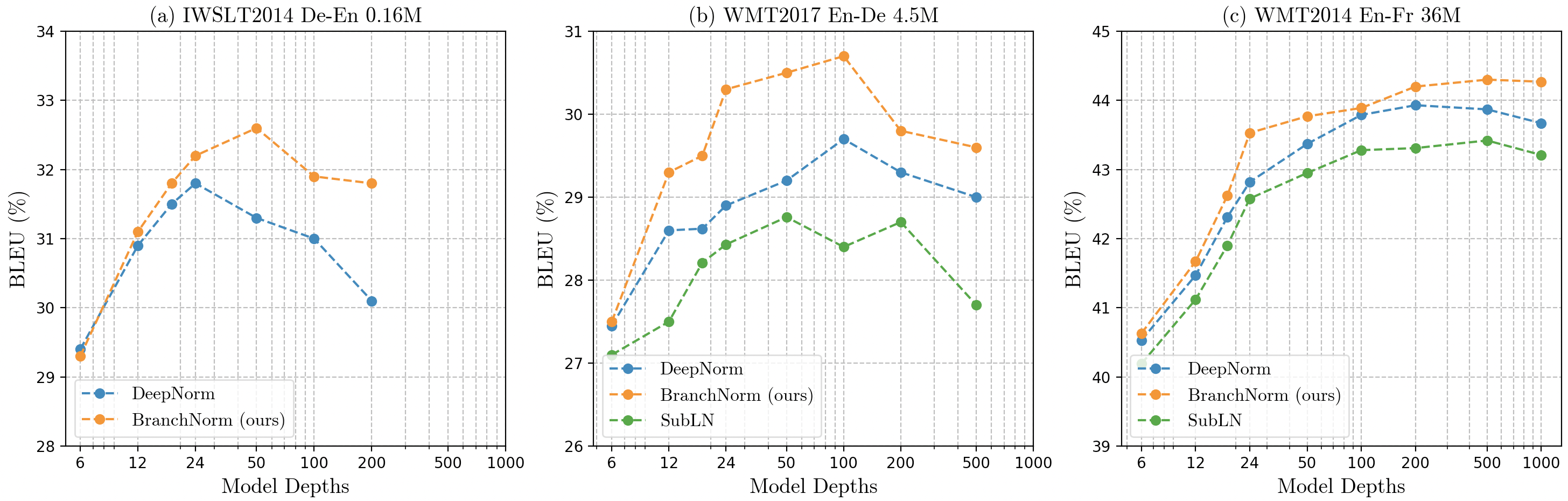}
      } 
      \caption{
      Performance of bilingual translation with different model depths, which are plotted on a logarithmic scale.
      % Note that the model depths are plotted on a logarithmic scale.
      } 
      \label{fig:bleu_curve}  
 \end{center} 
\end{figure*}
\subsection{Training Settings}
Our experiments are based on the fairseq code-base~\cite{fairseq}. For all experiments, we use the standard Transformer base setting which sets hidden dimensions to 512 and feed-forward inner representation to 2048, if not specifically noted. 
We initialize model parameters following DeepNorm~\citep{deepnet_2022}. 
All experiments are conducted on 32 NVIDIA A100 GPUs where each is allocated with a batch size of approximately 16,384 tokens.
All Transformer models are trained for 100k steps with the early stop for small-scale datasets.
The maximum norm step $T$ of BranchNorm in Equation~(\ref{equ:branchnorm_alpha}) is set to 4,000 for all experiments.
More details are elaborated in Appendix~\ref{appendix:hyper}.

\subsection{Bilingual Translation Tasks}
We compare several state-of-the-art approaches for deep Transformers, including DeepNorm~\cite{deepnet_2022}, Sub-LN~\cite{fundation_transformer}, NormFormer~\cite{Normformer2021}, ReZero~\cite{rezero2020} and {\em etc.}
% We re-implemented DeepNorm as it has no open-source code when we were conducting experiments.
We implemented DeepNorm and following the original paper~\citep{deepnet_2022} as the source codes were not publicly available when we conducted our experiments.
To ensure that the training framework is the same across different approaches, we followed the official source code of Sub-LN and NormFormer, and re-implemented them on Fairseq.
% Besides,  we reproduce the results of NormFormer and Sub-LN with their open-source codes.
Other results are directly cited from corresponding papers. 
\paragraph{Results on WMT17 En-De.}
Table \ref{tab:wmt17} reports the results of baselines and our approach on the WMT 2017 En-De dataset.
In most cases, the training of vanilla Post-LN Transformer get diverged due to its own training instability.
Meanwhile, previous approaches can stabilize the training of deep Transformer to varying degrees.
% And the Pre-LN Transformer is more stable for deep models and can be trained stably at 200 and 500 layers.
Results of our re-implemented DeepNorm slightly outperform those reported in the original paper, and serve as a stronger baseline to make the improvement of our approach more convincing.
It is noteworthy that all approaches show different degradation of BLEU score after 200 layers.
We preliminarily speculate that this phenomenon is caused by the overfitting on the small-scale WMT17 En-De after the model deepening.
In summary, our BranchNorm achieves the best results consistently at different depths and mitigates the performance degradation problem mentioned above
% Our BranchNorm achieves the best results with all the depths and has the least degradation with deeper models. 
Moreover, BranchNorm outperforms previous state-of-the-art deep models by up to +1.2 BLEU given the same model depths.
As shown in Figure~\ref{fig:train_step_bleu}, BranchNorm exhibits faster convergence and  better convergence performance than DeepNorm.

%%%%%%%%%%  OPUS table   %%%%%%%%%%
\begin{table*}[t]
\begin{center}
\scalebox{0.87}{
\begin{tabular}{l|cc|cccccc}
\toprule
\multirow{2}{*}{\textbf{Models}} & \multirow{2}{*}{\textbf{\# Layers}}  & \multirow{2}{*}{\textbf{\# Params}} & \multicolumn{3}{c}{OPUS100}  &  \multicolumn{3}{c}{MultiUN} \\
~ & ~ & ~ & \textbf{X$\rightarrow$En} & \textbf{En$\rightarrow$X} & \textbf{Avg} & \textbf{X$\rightarrow$En} & \textbf{En$\rightarrow$X} & \textbf{Avg} \\
\midrule
\multirow{3}{*}{Baseline~\citep{opus100}} & 12 & 133M &  27.5 & 21.4 & 24.5  & 43.8 & 52.3 &   48.1  \\
 & 24 & 173M & 29.5 & 22.9 & 26.2  & 46.1 & 53.9 &   50  \\
 & 48 & 254M & 31.4 & 24.0 & 27.7  & -- & -- & --  \\
 \midrule
\multirow{2}{*}{Pre-LN\cite{transformer_2017}} & 200 & 863M & 34.6 & 26.4 & 30.5  & 49.1 & 56.3 &  52.7 \\
 & 1000 & 3.8B & 34.0 & 28.0 & 31.0  & 50.1 & 56.7 & 53.4 \\
 \midrule
\multirow{2}{*}{DeepNorm\cite{deepnet_2022}} & 200 & 863M & 33.2 & \bf 29.0 & 31.1 & -- & -- & -- \\
 & 1000 & 3.8B & 33.9 & \bf 30.2 & 32.1 & -- & -- & --  \\
\multirow{2}{*}{DeepNorm $\dagger$ (\citeyear{deepnet_2022})} & 200 & 863M & 33.9 & 28.2 & 31.1 & 49.2 & 56.9 & 53.1  \\
 & 1000 & 3.8B & 34.8 & 29.4 & 32.1 & 50.3 & 57.2  &  53.8 \\
\midrule
\multirow{2}{*}{BranchNorm (ours)} & 200 & 863M & \textbf{34.2} & 28.5 & ~~\textbf{31.4}* & \bf 49.7 & 57.2 & ~~\textbf{53.4}* \\
 & 1000 & 3.8B & \textbf{35.0} & 29.6 & ~~\textbf{32.3}* & \bf 50.8 & \bf 57.6 &  ~~\textbf{54.2}* \\
\bottomrule
\end{tabular}
}
\caption{Average BLEU score(\%) of different models with varying  depths  on the OPUS-100 and MultiUN test sets. $\dagger$ indicates our reimplementations. The \textbf{bolded} scores correspond to the best in the same depths. `*' means BranchNorm is significantly better than DeepNorm with $p<0.05$.}
\label{tab:opus}
\end{center}
\end{table*}

\paragraph{Results on WMT14 En-Fr.}
Results of baselines and BranchNorm on the larger WMT 2014 En-Fr dataset are reported in Table \ref{tab:wmt14_enfr}. 
% The Vanilla Post-LN Transformer also fails in training when the model depth increase.
We observe similar findings with WMT 2014 En-De, namely, BranchNorm bring consistent improvements on models with different depths. Notably, our 500 layer model outperforms existing deep models and achieves a new SOTA performance of 44.3 BLEU.
% Moreover, we observe that larger scale data benefits more from model deepening, which is consistent with existing findings from large-scale pre-training models~\citep{scaling_laws_2022}.
% Unlike WMT 2014 En-De, the inflection point at which the model performance decreases with depth is delayed to 500 layers. Therefore, we inferred that more data require more depth which will be further discussed in Section \ref{sec:effects_of_data_scale}.
% BranchNorm still outperforms the previous deep mode up to +0.6 BLEU at all depths, and there is only a slight performance loss at 1000 layers.

\paragraph{Effects of Data Scale.}
% \label{sec:effects_of_data_scale}
We draw the detailed performance of three datasets with different scales in Figure~\ref{fig:bleu_curve}. 
Overall, we observe that as the model deepens, performance on smaller data is compromised, while larger datasets continue to benefit from the scaling of depth. 
This indicates that deeper models tend to require larger data to fit, which is consistent with findings on large-scale pretraining~\citep{scaling_laws_2022}.

\paragraph{Effects of Training Steps.}
In Figure~\ref{fig:train_step_bleu}, we plot the training curves of BranchNorm and DeepNorm for the 50L-50L model on the WMT2017 En-De.
The results demonstrate that BranchNorm can effectively unleash unleash the potential performance of deep models and finally yields a better converge performance.
In contrast, DeepNorm suffers from the undertraining problem, and performance is harmed to a certain extent at larger training steps. 

\iffalse
\subsection{Weighted-PreNorm}
PreNorm Transformer has very excellent stability compared to the Post-LN form, especially in the case of deep models. However, previous work claimed that with increasing depth, PreNorm degrades to increasing width instead of increasing depth and becomes significantly less effective. We also observe this problem in our experiments and significantly improve the performance of PreNorm Transformer under deep models by increasing the weights of the main branches. The calculation takes the following form:
\begin{equation}
x_{l+1} = \mathcal{F}(LN(x_l)) + \alpha x_l
\end{equation}
where $\alpha$ is a weight greater than 1. It is set to $Log(N)/4$ during our experiments.

\subsection{MixNorm Strategy}
To leverage the stability of the PreNorm and the superiority of the Post-LN, we propose a new structure called MixNorm. By adding two additional residual connections, we create a stable gradient stream on top of the Post-LN, improving the stability of the training while preserving the structure of the Post-LN. And no additional parameters are added. Each layer is calculated as follows:
\begin{equation}
x_{l+1} = \mathcal{F}(LN(x_l)) + LN(x_l) + x_l
\end{equation}
\fi

\begin{figure*}[t!]
\begin{center}
     \scalebox{0.8}{
      \includegraphics[width=1\textwidth]{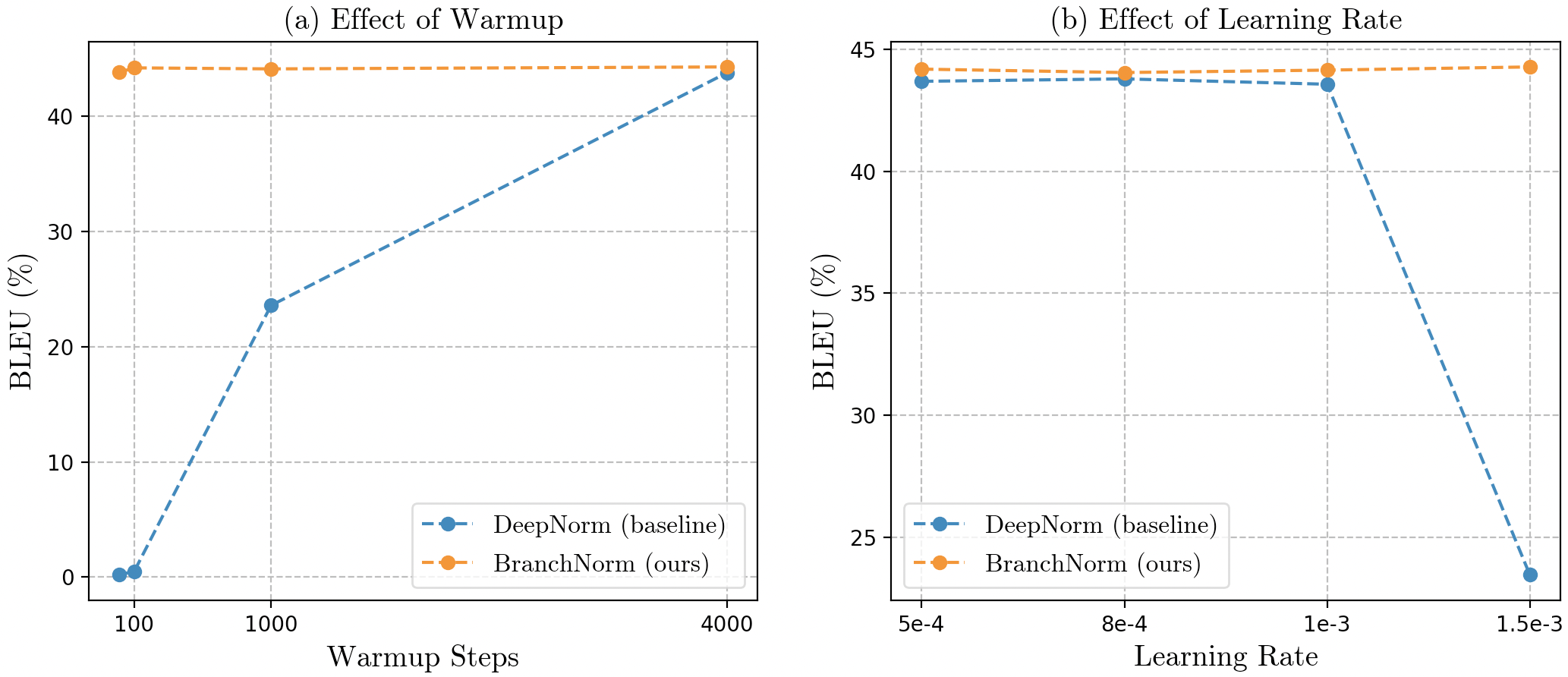}
      } 
      \caption{
      Effects of key hyperparameters ({\em i.e.,} warmup and learning rate) on training 100L-100L models on the WMT 2014 En-Fr dataset. 
      } 
      \label{fig:warmup_lr_diff}  
 \end{center} 
\end{figure*}

\begin{figure}[t!]
\begin{center}
     \scalebox{0.46}{
      \includegraphics[width=1\textwidth]{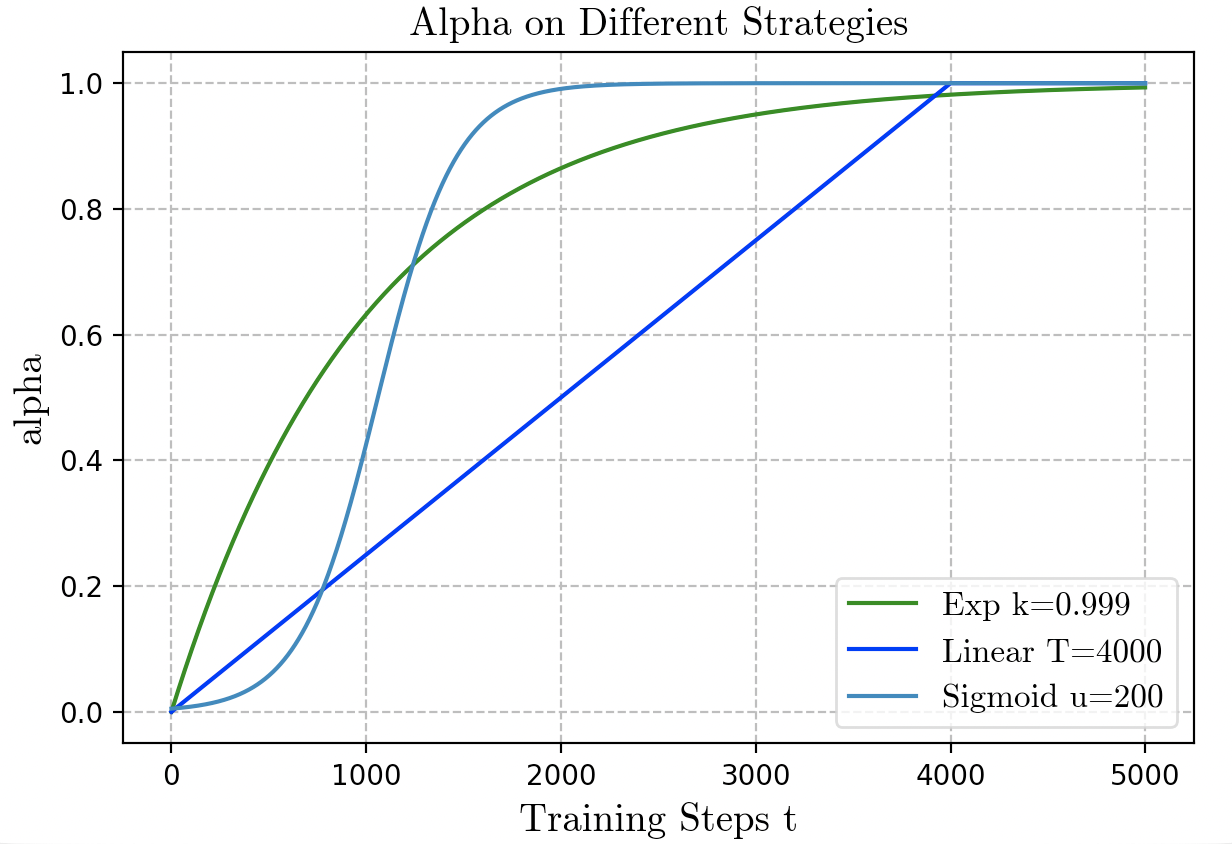} 
      } 
      \caption{
      Different growth strategies of $\alpha$ in BranchNorm. Note tht $\alpha$ is clipped to 1.0 for all strategies. 
      } 
      \label{fig:diffreent_alpha}  
 \end{center} 
\end{figure}

\begin{figure*}[t!]
\begin{center}
     \scalebox{0.85}{
      \includegraphics[width=1\textwidth]{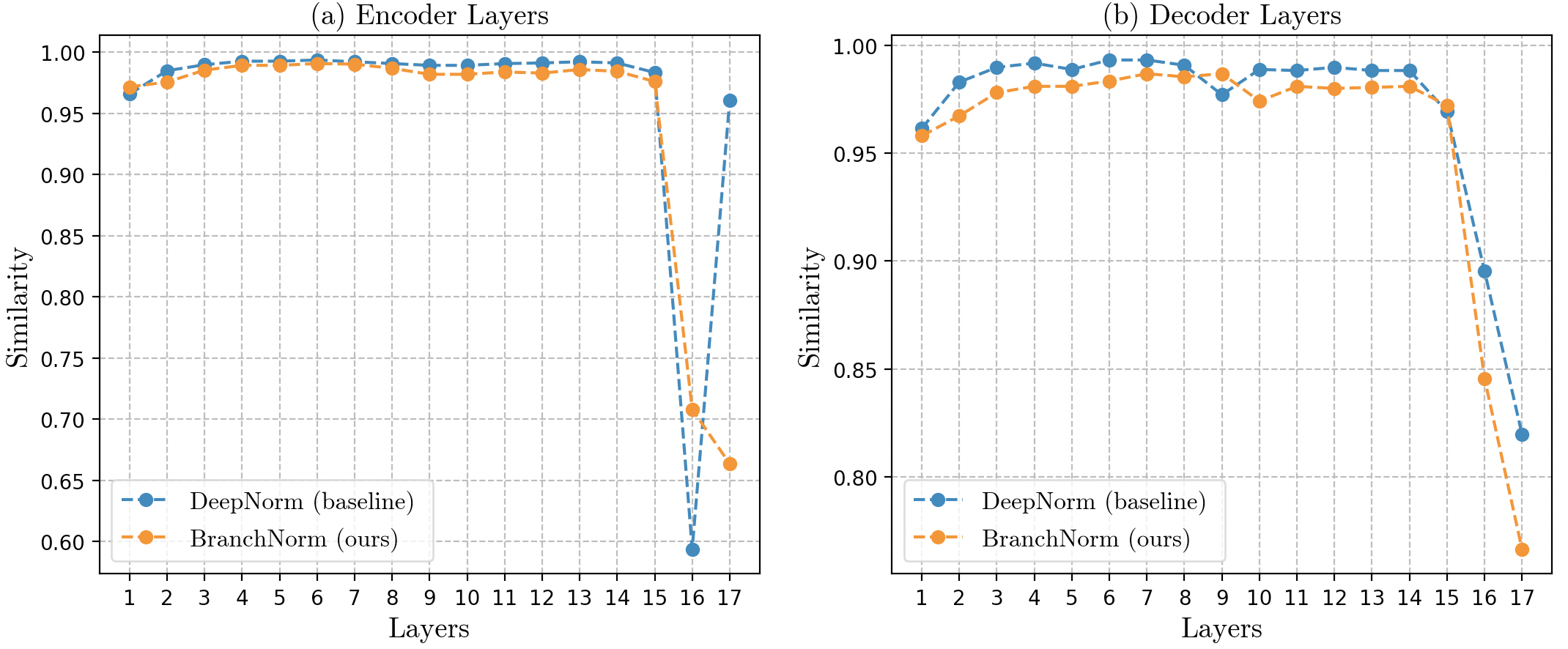}
      } 
      \caption{
      Representation similarity between adjacent layers. BranchNorm's values are lower than DeepNorm in most layers.
      } 
      \label{fig:layer_simi}  
 \end{center} 
\end{figure*}

\begin{figure*}[t!]
\begin{center}
     \scalebox{0.85}{
      \includegraphics[width=1\textwidth]{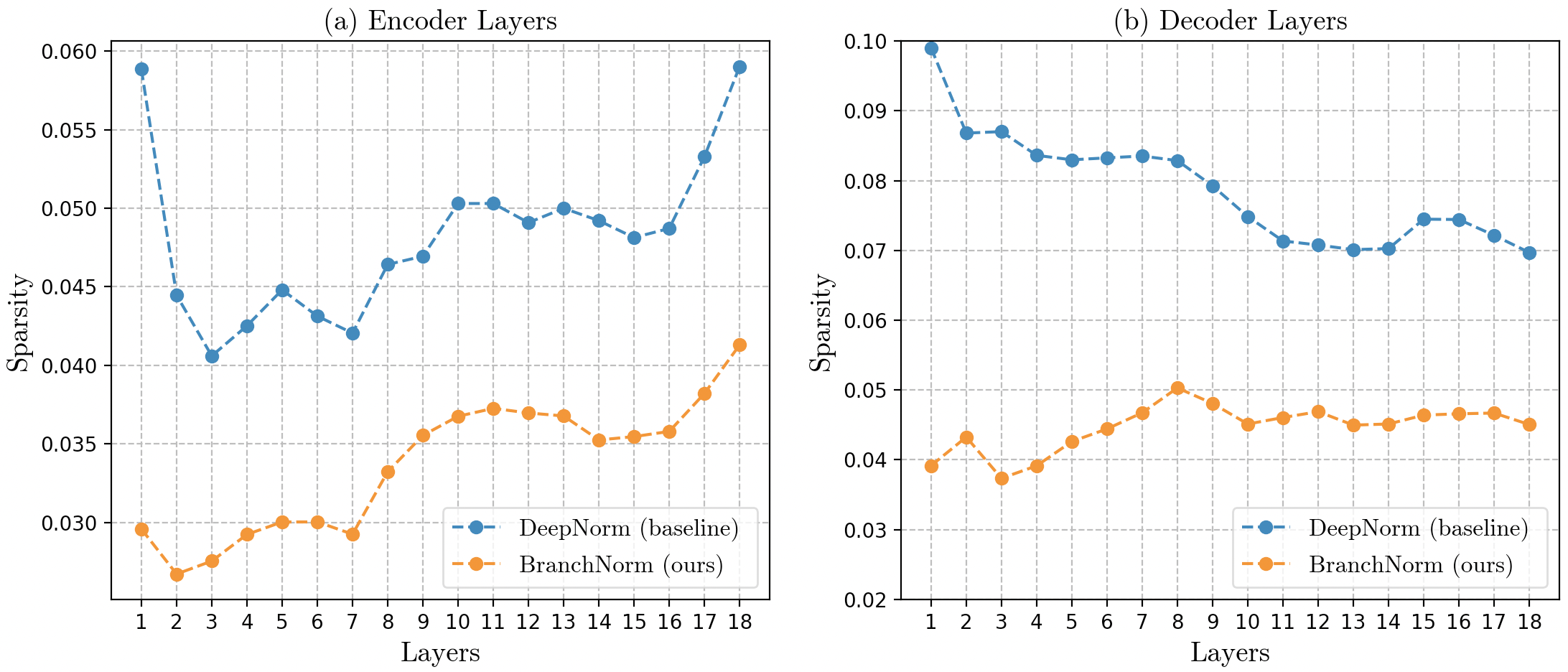}
      } 
      \caption{
      The sparsity of activation function of DeepNorm and BranchNorm models. The BranchNorm model is sparser than the DeepNorm one in all layers.
      } 
      \label{fig:sparsity}  
 \end{center} 
\end{figure*}

\subsection{Multilingual Translation Tasks}
% Multilingual translation tasks benefit more from layer scaling and perform more stably.
Results of various models on the OPUS-100 and MultiUN datasets are listed in Table \ref{tab:opus}.
% As the depth increases from 12 to 1000, the BLEU score increases by +7.8 on the OPUS dataset, by +6.1 on the MultiUN dataset, and by +2.6 on the WMT 2014 En-Fr dataset.
As the depth increases from 12 to 1000, the BLEU scores are increased by +7.8 and +6.1 points respectively on the OPUS dataset and MultiUN dataset.
Scaling the vanilla Pre-LN Transformer to 200 and 1000 layers proves to be ineffective, indicating that the vanilla Pre-LN Transformer is not effective enough on the deep model. 
BranchNorm consistently outperforms DeepNorm across all depths which is coherent with the conclusions of the bilingual translations.

\section{Analysis}
In this section, we first verify the robustness of BranchNorm to hyperparameter, and then analyze the parameter redundancy.

\subsection{Hyperparameter Sensitivity}
\label{sec:hyperpara_sensitivity}

\paragraph{Effects of Different $T$.}
We conduct experiments to evaluate the effect of varying the  different maximum norm step $T$ in Equation~(\ref{equ:branchnorm_alpha}) on BranchNorm.
A larger value of $T$ corresponds to a slower degradation of BranchNorm to the vanilla Post-LN, and generally yield  a more stable training process.
We vary $T \in [100, 400, 4000, 20000]$ and observe that BranchNorm is insensitive to the variation of $T$. 

\paragraph{Effects of Different Warmup and Learning Rate.}
We investigate the effect of these key hyperparameters on a 200-layer ({\em i.e.,} 100L-100L) Transformer on WMT14 En-Fr dataset and present the results in Figure~\ref{fig:warmup_lr_diff}.
Our observations indicate that BranchNorm is able to stably train a 200-layers Transformer without the use of warmup, and exhibits better tolerance for larger learning rates when compared to DeepNorm.

% \paragraph{Effects of different bias of $alpha$.}

\paragraph{Effects of Different Growing Strategies of $\alpha$.}
We investigate the effects of various growing strategies including the default linear strategy, which is illustrated in Figure~\ref{fig:diffreent_alpha}. 
% Preliminary experimental results of 200-layer Transformers on WMT14 En-Fr indicate that our method 
For 200-layer Transformers on WMT14 En-Fr,  we respectively obtain 44.20, 44.15, and 44.21 BLEU on the linear, exp, and sigmoid strategies, indicating that our method is robust to these strategy variants, therefore, we employ the simplest linear strategy in all experiments.

\iffalse
\subsection{Noise Resistance}

Adding different level of noisy data to cause unbanlanced gradient for deep models.
Too much noise in turn leads to stable model training, presumably due to its stronger normalization effect.
\fi

\subsection{Parameter Redundancy}
% Previous studies~\citep{LiyuanLiu2020OnTV} claim that Pre-LN Transformer has relatively larger weights for its residual branch which in turn limits its potential performance as the model deepens. 
% We suspect that DeepNorm directly increases the weights of the residual branches to improve the stability of deep model training, but may limit the potential of the deep model as well.
% To verify our assumption, we computed the similarity of representations between adjacent layers from the models trained with different approaches.
% We train two models that have 36 layers with DeepNorm and BranchNorm. 
\paragraph{Representation Similarity.} 
Previous studies~\citep{LiyuanLiu2020OnTV} has posited that the Pre-LN Transformer has disproportionately large weights on its residual branch, which may inhibit its potential performance as the model deepens.
Our hypothesis is that DeepNorm directly augment the weights of the residual branches in order to enhance the stability of deep model training, but may also impede the potential of the deep model.
In order to verify this assumption, we employed a methodology to determine the cosine similarity of representations between adjacent layers in 200-layer ({\em i.e.,} 100L-100L)  models that were respectively trained with DeepNorm and BranchNorm.

% Figure \ref{fig:layer_simi} shows the representation similarity of both encoder and decoder layers. The similarity score of DeepNorm is always higher than BranchNorm which means increasing the weights of the residual branch causes the model to be more like the Pre-LN Transformer. Then, this property in turn may lead to the degradation of the deep model and depresses the performance. Therefore, it is essential to get the weights back to the vanilla states. 

The representation similarity of both encoder and decoder layers is presented in Figure~\ref{fig:layer_simi}. It is observed that the similarity score of DeepNorm consistently exceeds that of BranchNorm, indicating that the augmentation of the weights of the residual branch results in the model becoming more akin to the Pre-LN Transformer.
Similar findings about sparsity are consistently observed for models with different depths and data.
This characteristic may subsequently contribute to the degradation of the deep model and negatively impact performance.
Therefore, it is essential to revert the weights to their original values, as is done in the implementation of BranchNorm.

\paragraph{Sparsity of Activation Function.} 
% Another characteristic of the Transformer that has been recently discussed by researchers is sparsity. 
\citet{li2022large} studies the activation function sparsity of the Transformer and demonstrates that the sparser model comes with better generalization and robustness. The sparsity is quantified by the percentage of nonzero entries after the activation function. As shown in Figure \ref{fig:sparsity}, we observe the sparsity of two models trained with DeepNorm and BranchNorm respectively, and find that BranchNorm had a relatively smaller sparsity. 
% Figure \ref{fig:sparsity} shows the sparsity result of the activation function. Both models have a very low sparsity(e.g., 6.0\% for the encoder and 10\% for the decoder part), and BranchNorm has lower values at any layer.
To confirm the effect of sparsity on the robustness and generalization of the model, we conducted further experiments on the MTNT~\cite{michel-neubig-2018-mtnt}. MTNT (Machine Translation of Noisy Text) consists of noisy comments on Reddit (www.reddit.com) and professionally sourced translations and is a testbed for robust translation.
We evaluate two En-Fr models that are trained with DeepNorm and BranchNorm on this noise dataset. BranchNorm has a significant improvement of 1.0 BLEU over DeepNorm, indicating that our method is able to improve robustness by increasing the sparsity of the model.

\section{Conclusion}
In this paper, we first explore the undertraining problem of DeepNorm and propose a more flexible canonical approach, namely BranchNorm, which theoretically stabilizes the training with smooth gradient norms at the early stage.
Once the dangerous phase of training instability is passed, BranchNorm can then degenerate to a standard Post-LN, thus encouraging better convergence performance.
Experiment results on several translation tasks show that BranchNorm achieves a better trade-off between training stability and converge performance.

\section*{Limitations}
The training of deep Transformers generally requires large GPU resources, for example, training a 1,000-layer WMT14 En-Fr translation model requires 1000 GPU days.
In addition, deeper decoders can lead to slower inference, and more model architecture design or compression techniques need to be further explored to make deep models practically deployable for applications.

% \section*{Ethics Statement}
% Scientific work published at ACL 2023 must comply with the ACL Ethics Policy.\footnote{\url{https://www.aclweb.org/portal/content/acl-code-ethics}} We encourage all authors to include an explicit ethics statement on the broader impact of the work, or other ethical considerations after the conclusion but before the references. The ethics statement will not count toward the page limit (8 pages for long, 4 pages for short papers).

\bibliography{anthology,custom}
\bibliographystyle{acl_natbib}

% \newpage
\appendix

\iffalse
\section{Examples of Pre-Norm and Post-Norm}
An example of these two architectures is shown in Figure~\ref{fig:prenorm_Post-LN}.
\begin{figure}[t!]
\begin{center}
     \scalebox{0.45}{
      \includegraphics[width=1\textwidth]{figures/prenorm-postnorm.pdf}
      } 
      \caption{
      The architectures of Pre-Norm (Pre-LN) and Post-Norm (Post-LN) Transformers.
      } 
      \label{fig:prenorm_Post-LN}  
 \end{center} 
\end{figure}
\fi

\section{Theoretical Proofs}  
\label{appendix:proof}

\subsection{Gradients of Post-LN}
\label{appendix:proof_postln}
Given a Transformer with $L$ sub-layers and the training loss $\mathcal{E}$, the gradient for the $l$-th sub-layer is calculated by the chain rule:

\begin{equation}
\begin{aligned}
\frac{\partial \mathcal{E}}{\partial x_l}=
\frac{\partial \mathcal{E}}{\partial x_L}
\frac{\partial x_L}{\partial x_l}
\end{aligned}
\label{equ:proof_polstln}
\end{equation}
Recursively decomposing $\frac{\partial x_L}{\partial x_l}$ in the above equation, we have:
\begin{equation}
\begin{aligned}
\frac{\partial x_L}{\partial x_l}=
\frac{\partial x_L}{\partial x_{L-1}}
\frac{\partial x_{L-1}}{\partial x_{L-2}}
\cdots
\frac{\partial x_{l+1}}{\partial x_l}
\end{aligned}
\label{equ:proof_chain_rule}
\end{equation}
Given the Post-LN calculate the $x_{l+1}$ as :
\begin{equation}
x_{l+1} = LN(x_l + \mathcal{F}\left(x_l ; \theta_l\right))    
\end{equation}
If we name the output of residual connection as $y_l = x_l + \mathcal{F}\left(x_l ; \theta_l\right)$, we can calculate the partial derivatives of two adjacent layers as:
\begin{equation}
\begin{aligned}
\frac{\partial x_{l+1}}{\partial x_l}&=
\frac{\partial x_{l+1}}{\partial y_l}
\frac{\partial y_{l}}{\partial x_l}\\
&= \frac{\partial \mathrm{LN}\left(y_l\right)}{\partial y_l}\left( 1 + \frac{\partial \mathcal{F}\left(x_l ; \theta_l\right) }{\partial x_l}  \right)
\end{aligned}
\label{equ:proof_adjancent}
\end{equation}
We put Equation~(\ref{equ:proof_adjancent}) and Equation~(\ref{equ:proof_chain_rule}) into Equation~(\ref{equ:proof_polstln}) and get:
\begin{equation}
\begin{aligned}
\frac{\partial \mathcal{E}}{\partial x_l}=
&\underbrace{\frac{\partial \mathcal{E}}{\partial x_L}}_{irreducible}
\times 
\underbrace{\prod_{k=l}^{L-1} \frac{\partial \mathrm{LN}\left(y_k\right)}{\partial y_k}}_{LN}
\times \\
&\underbrace{\prod_{k=l}^{L-1}\left(1+\frac{\partial \mathcal{F}\left(x_k ; \theta_k\right)}{\partial x_k}\right)}_{residual}
\end{aligned}
\label{equ:proof_grad_postln}
\end{equation}
the above gradient consists of three terms and the last two items are multiplications with respect to the number of model layers $L$.
Once $L$ get larger, the gradient of Post-LN will face the risk of vanishing or exploding.

\iffalse
\subsection{Change in Gradient of LayerNorm with Input Scaling}
In this section, we aims to analyze how the gradient of the LayerNorm $\frac{\partial \mathrm{LN} (x) }{\partial x}$ changes when the input $x$ scaling into $\alpha x$, where $\alpha > 1$ is a scalar multiplier.
Formally, LayerNorm normalize the input by:
\begin{equation}
    y = \mathrm{LN}({x}) = \frac{\gamma ({x} - \mu)}{\sigma}  + \beta
\end{equation}
where $\mu$ and $\sigma$ are the mean and standard deviation, respectively, and $\gamma$ and $\beta$ are learnable parameters.
When we scale $x$ into $\alpha x$, the output of LayerNorm changes as follows:
\fi

\subsection{Gradients of DeepNorm}
\label{appendix:proof_deepnorm}

DeepNorm rescales the residual branch with a scalar multiplier $\alpha > 1$, and calculates the sub-layer as follows:
\begin{equation}
    x_{l+1} = LN(\alpha x_l + \mathcal{F}\left(x_l ; \theta_l\right))
\end{equation}
Follow the above process in~\ref{appendix:proof_postln}, we have the gradient of DeepNorm:

\begin{equation}
\resizebox{1.0\linewidth}{!}{
$
\begin{aligned}
\frac{\partial \mathcal{E}}{\partial x_l}&=
\underbrace{\frac{\partial \mathcal{E}}{\partial x_L}}_{irreducible}
\times 
\underbrace{\prod_{k=l}^{L-1} \left( \frac{\partial \mathrm{LN}\left( \alpha x_k + \mathcal{F}\left(x_k ; \theta_l\right)  \right)}{\partial \left( \alpha  x_k + \mathcal{F}\left(x_k ; \theta_l\right) \right)} \right) }_{LN} \times  \\
& \ \ \ \ \ \ \ \ \ \underbrace{\prod_{k=l}^{L-1}\left(\alpha + \frac{\partial \mathcal{F}\left(x_k ; \theta_k\right)}{\partial x_k} \right) }_{residual} \\
\end{aligned} 
$
}
\end{equation}
Given that DeepNorm assigns a relative larger value for $\alpha$ to make it to amplify the output percentage of residual connections.
Here, we introduce an assumption to simplify the derivation: If $\alpha$ gets large enough, we can approximate the above equation as follows:
\begin{equation}
\resizebox{1.0\linewidth}{!}{
$
\begin{aligned}
\frac{\partial \mathcal{E}}{\partial x_l}& \approx 
\underbrace{\frac{\partial \mathcal{E}}{\partial x_L}}_{irreducible}
\times 
\underbrace{\prod_{k=l}^{L-1} \left( \frac{\partial \mathrm{LN}\left( \alpha x_k  \right)}{\partial \left( \alpha  x_k  \right)} \right) }_{LN} 
\times 
\underbrace{\prod_{k=l}^{L-1} \alpha }_{residual} 
\end{aligned} 
$
}
\end{equation}
We let $z_k = \alpha x_k $ and use the chain rule, then get:

\begin{equation}
\resizebox{1.0\linewidth}{!}{
$
\begin{aligned}
\frac{\partial \mathcal{E}}{\partial x_l}& = 
\underbrace{\frac{\partial \mathcal{E}}{\partial x_L}}_{irreducible}
\times 
\underbrace{\prod_{k=l}^{L-1} \left( \frac{\partial \mathrm{LN}\left( z_k  \right)} {\partial x_k} \times 
\frac{\partial x_k } {\partial z_k}  \right) }_{LN}
\times 
\underbrace{\prod_{k=l}^{L-1} \alpha }_{residual} \\
&=\underbrace{\frac{\partial \mathcal{E}}{\partial x_L}}_{irreducible}
\times 
\underbrace{\prod_{k=l}^{L-1} \left( \frac{\partial \mathrm{LN}\left(x_k  \right)} {\partial x_k} \times 
\frac{1} {\alpha}  \right) }_{LN} 
\times 
\underbrace{\prod_{k=l}^{L-1} \alpha }_{residual} \\
&=\underbrace{\frac{\partial \mathcal{E}}{\partial x_L}}_{irreducible}
\times 
\underbrace{\prod_{k=l}^{L-1} \frac{\partial \mathrm{LN}\left(x_k  \right)} {\partial x_k}  }_{LN} 
\end{aligned} 
$
}
\end{equation}
When compared with the gradient of Post-LN in Equation (\ref{equ:proof_grad_postln}), DeepNorm can approximately eliminate the final multiplication item, and thus mitigate the risk of gradient vanishing or exploding to a certain degree. 

\subsection{Gradients of BranchNorm}
BranchNorm directly rescale the non-residual branch in Transformer and conduct calculations for the $l$-th sub-layer as:
\begin{equation}
    x_{l+1} = LN(x_l + \alpha 
 \mathcal{F}\left(x_l ; \theta_l\right))
\end{equation}
Similar to the previous analysis process, we can calculate the gradients of BranchNorm as:
\begin{equation}
\begin{aligned}
\frac{\partial \mathcal{E}}{\partial x_l}&=
\underbrace{\frac{\partial \mathcal{E}}{\partial x_L}}_{irreducible}
\times 
\underbrace{\prod_{k=l}^{L-1}  \frac{\partial \mathrm{LN}\left(x_k + \alpha \mathcal{F}\left(x_k ; \theta_l\right) \right)}{\partial \left( x_k + \alpha \mathcal{F}\left(x_k ; \theta_l\right) \right)} }_{LN} 
\times \\ 
&\ \ \ \ \ \ \ \ \underbrace{\prod_{k=l}^{L-1} \left(1 + \alpha \frac{\partial \mathcal{F}\left(x_k ; \theta_l\right)  }{\partial x_k} \right)
 }_{residual}\\
&=\underbrace{\frac{\partial \mathcal{E}}{\partial x_L}}_{irreducible}
\times 
\underbrace{\prod_{k=l}^{L-1} \frac{\partial \mathrm{LN}\left(x_k\right)}{\partial x_k} }_{LN}  \ \ \ \ \ (\alpha = 0)
\\
\end{aligned} 
\end{equation}
\iffalse
\begin{equation}
\begin{aligned}
\frac{\partial \mathcal{E}}{\partial x_l}&=
\underbrace{\frac{\partial \mathcal{E}}{\partial x_L}}_{irreducible}
\times 
\underbrace{\prod_{k=l}^{L-1}  \frac{\partial \mathrm{LN}\left(x_k\right)}{\partial x_k} }_{LN}
\times 
\underbrace{\prod_{k=l}^{L-1}1}_{residual}\\
&=\underbrace{\frac{\partial \mathcal{E}}{\partial x_L}}_{irreducible}
\times 
\underbrace{\prod_{k=l}^{L-1} \frac{\partial \mathrm{LN}\left(x_k\right)}{\partial x_k} }_{LN}  \ \ \ \ \ (\alpha = 0)
\\
\end{aligned} 
\end{equation}
\fi
BranchNorm can stabilize the gradient norm into while DeepNorm require a relatively strong assumption in Equation~(\ref{equ:grad_deepnorm_appro}). Experimentally, in Figure~\ref{fig:grad_loss}, we observe corresponding  smoother gradients of BranchNorm at the very beginning of training.

\section{Hyperparameter} 
\label{appendix:hyper}

%%%%%%%%%%%%%%%%%%%%%%%%%%%%%%%%%%%%%%%%%%%%%%%%%%%%%%%%%%%%%%%%%%%%%%%%%%%%%%%%
\begin{table*}[htbp]
\begin{center}
\begin{tabular}{l|ccc}
\toprule
\textbf{Hyperparameters} &  \textbf{Small Scale} & \textbf{Medium Scale} &  \textbf{Large Scale} \\
\midrule
Learning rate & \multicolumn{3}{c}{5e-4}  \\
Learning rate scheduler & \multicolumn{3}{c}{inverse sqrt} \\
Warm-up updates & \multicolumn{3}{c}{4000} \\
Warm-up init learning rate & \multicolumn{3}{c}{1e-7} \\
Max tokens & \multicolumn{3}{c}{128 $\times$ 4096} \\
Adam $\epsilon$ & \multicolumn{3}{c}{1e-8} \\
Adam $\beta$ & \multicolumn{3}{c}{(0.9, 0.98)} \\
Label smoothing & \multicolumn{3}{c}{0.1} \\
Training updates & \multicolumn{3}{c}{100K} \\
\midrule
Gradient clipping & \multicolumn{3}{c}{0.0} \\
Dropout & 0.4 & 0.2 & 0.1 \\
Weight decay & \multicolumn{3}{c}{0.0001} \\
\midrule
Hidden size & \multicolumn{3}{c}{512} \\
FFN inner hidden size & \multicolumn{3}{c}{2048} \\
Attention heads & \multicolumn{3}{c}{8} \\
\bottomrule
\end{tabular}
\caption{Hyperparameters for the Transformer$_{base}$ experiments on different data sizes. `Small Scale': IWSLT 2014 De-En and WMT17 En-De. `Medium Scale': WMT14 En-Fr.
`Large Scale': OPUS-100 and MultiUN datasets.}
\end{center}
\end{table*}

%%%%%%%%%%%%%%%%%%%%%%%%%%%%%%%%%%%%%%%%%%%%%%%%%%%%%%%%%%%%%%%%%%%%%%%%%%%%%%%%

\end{document}